**An Innovative Attack Modelling and Attack Detection Approach for a Waiting Time-based Adaptive Traffic Signal Controller**

**Sagar Dasgupta***
Ph.D. Student
Department of Civil, Construction & Environmental Engineering
The University of Alabama
3014 Cyber Hall, Box 870205, 248 Kirkbride Lane, Tuscaloosa, AL 35487
Tel: (864) 624-6210; Email: sdasgupta@crimson.ua.edu

**Courtland Hollis**
Undergraduate Student
Department of Civil, Construction & Environmental Engineering
The University of Alabama
3014 Cyber Hall, Box 870205, 248 Kirkbride Lane, Tuscaloosa, AL 35487
Tel: 678-245-9036; Email: cmhollis2@crimson.ua.edu

**Mizanur Rahman, Ph.D.**
Assistant Professor
Department of Civil, Construction & Environmental Engineering
The University of Alabama
3015 Cyber Hall, Box 870205, 248 Kirkbride Lane, Tuscaloosa, AL 35487
Tel: (205) 348-1717; Email: mizan.rahman@ua.edu

**Travis Atkison, Ph.D.**
Assistant Professor
Department of Computer Science
The University of Alabama
3057 Cyber Hall, Box 870290, Tuscaloosa, AL 35487
Tel: (205) 348-4740; Email: atkison@cs.ua.edu

*Corresponding author

**ABSTRACT**
An adaptive traffic signal controller (ATSC) combined with a connected vehicle (CV) concept uses real-time vehicle trajectory data to regulate green time and has the ability to reduce intersection waiting time significantly and thereby improve travel time in a signalized corridor. However, the CV-based ATSC increases the size of the surface vulnerable to potential cyber-attack, allowing an attacker to generate disastrous traffic congestion in a roadway network. An attacker can congest a route by generating fake vehicles by maintaining traffic and car-following rules at a slow rate so that the signal timing and phase changes without having any abrupt changes in number of vehicles. Because of the adaptive nature of ATSC, it is a challenge to model this kind of attack and also to develop a strategy for detection. This paper introduces an innovative "slow poisoning" cyberattack for a waiting time based ATSC algorithm and a corresponding detection strategy. Thus, the objectives of this paper are to: (i) develop a "slow poisoning" attack generation strategy for an ATSC, and (ii) develop a prediction-based "slow poisoning" attack detection strategy using a recurrent neural network—i.e., long short-term memory model. We have generated a "slow poisoning" attack modeling strategy using a microscopic traffic simulator—Simulation of Urban Mobility (SUMO)—and used generated data from the simulation to develop both the attack model and detection model. Our analyses revealed that the attack strategy is effective in creating a congestion in an approach and detection strategy is able to flag the attack.

**Keywords:** Cybersecurity, slow poisoning attack, adaptive traffic signal controller, long short-term memory (LSTM)

**INTRODUCTION**

Rapidly developing technologies are radically changing transportation systems. However, the evolution of mainstream transportation systems towards a connected cyber infrastructure, such as connected traffic signal controllers, is increasing system vulnerability to potential cyber attack, allowing malicious actors (individuals, criminals, or terrorist organizations) to exploit security vulnerabilities of such transportation infrastructure *(1)-(3)*. In the U.S., the number of cyber-attacks on smart mobility systems has jumped significantly in recent years *(4)*. As vehicles are moving towards connected and automated driving, and cities are focusing on creating a transportation cyber infrastructure that will transform legacy transportation infrastructure to connected, adaptable, and automated systems, the security problems will only increase and further compromise public safety *(5)*. Many studies show that a cyber attack on connected vehicle-based (CV-based) traffic signal control algorithms can break down a traffic network by creating severe congestion *(6-10)*.

An adaptive traffic signal controller (ATSC) combined with a connected vehicle (CV) concept uses real-time vehicle trajectory data to regulate green time; this combination also has the ability to reduce intersection waiting time significantly and improve travel time in a signalized corridor *(11)*. A CV-based ATSC can be manipulated in two ways: (i) gain access through vulnerabilities and exploit the ATSC; and (ii) produce abnormal behavior through the manipulation of inputs of vehicle-related data *(9)*. This study focuses on the latter type of manipulation. An attacker can compromise the timing and phase plan generation function by generating falsified vehicle trajectory data. If a traffic signal controller were victim to such a cyberattack, there could be a significant delay at a signalized intersection due to traffic congestion caused by the inaccurate function of the traffic signal control algorithm *(6-10)*.

To date, different types of cyber attack models have been developed to assess the impact of the attack on the traffic signal control algorithm and traffic network. Yen et al. *(12)* developed a time spoofing attack and evaluated the impact of the attack on four different backpressure-based control algorithms—i.e., the delay-based, queue-based, sum-of-delay-based, and a hybrid scheme that combines delay-based and queue-based schemes. A time spoofing attack usually alters and increases the arrival times of vehicles towards a roadway intersection. In another study, Yen et al. *(13)* created two types of attacks: (i) a time spoofing attack and (ii) a ghost vehicle attack. In a ghost vehicle attack, a vehicle can hide itself by being disconnected from traffic signal control and eventually affect the signal control algorithm. Later, Yen et al. evaluated the impact of these attacks on a backpressure-based algorithm for traffic signal controller. Furthermore, they presented two algorithms—auction-based and hybrid-based—to mitigate the impact of time spoofing and ghost vehicle attack *(13)*. In another study, Chen et al. *(9)* created a data spoofing attack and analyzed the vulnerability of the attack. Recently, Comert et al. *(10)* modeled a belief-network based attack to evaluate the vulnerability of a traffic network. However, to the best of the authors' knowledge, there is no research on attack modeling and detection related to ATSC when an attacker starts slowly injecting fake vehicles following traffic and car-following rules. It is a challenge to differentiate between a cyber-attack and a variation of the traffic pattern because of the dynamic nature of vehicle arrival rates at the intersection. In addition, the adaptive nature of ATSC makes it more complex to model this type of attack as well as to develop a strategy to detect such attack.

This paper introduces an innovative "slow poisoning" cyberattack for a waiting time-based ATSC and corresponding detection strategy. Thus, the objectives of this paper are: (i) develop a "slow poisoning" attack generation strategy for a waiting time-based ATSC; and (ii) develop a prediction-based "slow poisoning" attack detection strategy using a recurrent neural network—i.e., long short-term memory (LSTM) model *(15)*.

**RELATED WORK**
As a traffic controller is a hardware-based failsafe, vulnerabilities of a controller lead to unexpected and disastrous traffic congestion instead of accidents. An attacker can gain access to a signal controller through wired or wireless connectivity and can compromise the system in two different ways: (i) by taking control of the signal controller and changing signal timing and eventually controlling the signal phases; and (ii) by generating false vehicle trajectory data to change signal timing. By tampering a traffic signal, an attacker can even take control of a transportation network. Laszka et al. developed an evaluation approach for analyzing the vulnerabilities of a transportation network and identifying critical traffic signals that can have maximum impact on the roadway network and generate congestion *(9)*. Through numerical and simulation experiments, they evaluated both randomly generated and real-world networks and proved that their approach is efficient in finding network vulnerabilities and corresponding critical signals. Later, Perrine et al. studied a district-wide cyber attack on traffic signals and conducted a risk assessment of the attack *(7)*. They used a dynamic traffic assignment network to model a central business district and evaluate the impact of a cyber attack by selectively disabling traffic signals. The travel delay at the intersection was found to be 4.3 times when they disable 26 signals. However, a similar impact can be achieved by only disabling seven traffic signals if it is possible to maximize the number of travelers passing through those intersections. Recently, Thodi et al. developed a bi-objective optimization model to maximize the impact of a cyber attack on the network by increasing traffic congestion and minimizing changes in traffic signal timing *(8)*. They investigated an urban traffic network's vulnerability to cyber attacks on traffic signals and found that roadway network vulnerability does not depend on traffic demands. In addition, they concluded minor changes in a traffic signal could create disastrous traffic congestion.

    Besides the foregoing literature related to network vulnerability, there have been several studies that investigated how a cyberattack can be generated/modeled and mitigated. Yen et al. evaluated four different backpressure-based control algorithms for a signal controller—i.e., the delay-based, queue-based, sum-of-delay-based, and a hybrid scheme that combines delay-based and queue-based schemes *(12)*. A time spoofing attack affects the transportation system by disrupting the time phasing of a traffic signal controller by altering arrival times of vehicles towards a roadway intersection. They found that the delay-based scheme is more vulnerable to time spoofing attacks compared to sum-of-delay-based schemes. However, delay-based shows better performance in terms of the fairness index related to delay performance. Later, Yen et al. evaluated the impact of a time spoofing attack and ghost vehicle attack (a ghost vehicle refers to a vehicle that can hide itself by being disconnected from traffic signal control) on backpressure-based algorithm for traffic signal control *(13)*. They found that these attacks could influence backpressure-based signal phases. In addition, they presented two algorithms—auction-based and hybrid-based—to mitigate the impact of time spoofing and ghost vehicle attacks. In *(9)*, Chen et al. analyzed the system design of a traffic signal controller and how a data spoofing attack can be generated on it. They showed that current traffic signal control algorithms are vulnerable to a data spoofing attack even with a single vehicle. Such attacks can reduce the mobility benefit of CV-based signal control systems by 23%. In another study, Feng et al. identified four attack surfaces for CV-based traffic signal control systems: (i) signal controllers, (ii) vehicle detectors, (iii) roadside units, and (iv) onboard units. Their empirical study showed that the travel delay on the

different attack scenarios, and an attack may even improve the system performance if it is not effective *(14)*. Even critical signals in real-world network were identified through simulation evaluation for a signalized roadway network. In another study, Comert et al. developed a belief-network based attack modeling for CV-based signalized networks *(10)*. They have assessed the risk of the attack model using simulation in terms of queue length and found that there is an increase of 15% and 125% queue with and without redundant systems, respectively.

Although several studies exist related to attack modeling and evaluating the imapct of cyberattacks on CV-based traffic signal controller, no study exists on "slow poisoning" attacks on an ATSC that can fool a prediction-based attack detection strategy. Thus, we develop a "slow poisoning" attack modelling strategy and an attack detection strategy to flag the attack.

**THREAT MODEL**

For a CV-based ATSC, vulnerabilities in the communication network or traffic signal controller itself allow modification of traffic control related data. We assume that an attacker can generate fake vehicles with fake positions, which approach to a subject intersection (see **Figure 1**). A CV-based ATSC receives broadcasted basic safety messages (BSMs) from generated fake vehicles, and we also assume that fake vehicles are generated in such a way so that all the fake vehicles follow traffic safety and safe gap as per car-following rules. The ATSC receives fake broadcasted BSMs and controls the signal phasing and timing accordingly. Before implementing any attack, an attacker must perform a survey so that he/she knows the types of signal controller algorithms and can figure out the corresponding vulnerabilities, signal control configurations, geometry of intersection, and lane configuration.

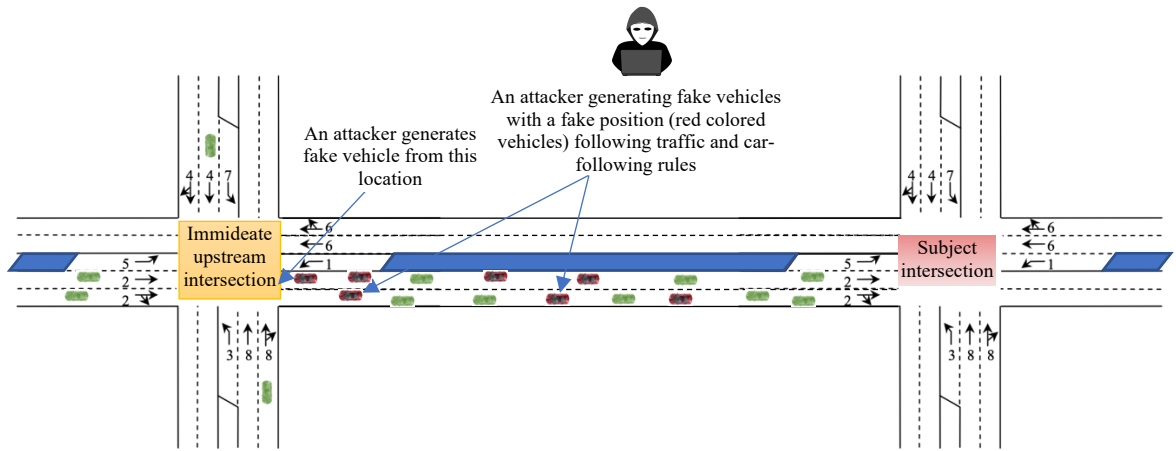

**Figure 1 An example scenario of "slow poisoning" attack following traffic and car-following rules**

To make the attack scenario more realistic, we assume that the attacker stays close to the subject intersection and uses the required devices, such as a laptop and BSM broadcasting device. Using a parallel computing option, an attacker can create multiple fake vehicles and generate BSMs. In addition, we assume that all vehicles approaching a subject intersection could exchange vehicle waiting time and vehicle trajectory data with the controllers through wireless connectivity. The ATSC decides signal phase and timing with the highest average waiting time between all movements (see more details of a waiting time based ATSC in the "Attack Modelling" section).

*Dasgupta, Hollis, Rahman, and Atkison*

Furthermore, we also assume that the machine learning based cyber attack detection model could detect attacks by predicting the number of vehicles arriving at the subject intersection. However, a machine learning model would be vulnerable if an attacker injects vehicles on the east bound (EB) approach (or any other approach) of the subject intersection with a slow vehicle generation rate and follows roadway safety rules. The attacker can generate fake vehicles in one or multiple approaches of the subject intersection at the same time. As the ultimate goal of the attacker is to create congestion by manipulating the vehicle data by slowly introducing fake vehicles and thus by influencing the signal controller decision, a time-series based machine learning model might not be able to distinguish between the number and waiting time information of fake and actual vehicles created by an attacker.

**ATTACK MODELLING**

We have designed an attack model by generating fake vehicles and the corresponding trajectory data using the Simulation of Urban Mobility (SUMO) microscopic traffic simulator. This section details attack goals, simulation set-up, ATSC algorithm and data generation, and justification for achieving attack goals. We also present attack-free and attack data in the "ATSC algorithm and data generation" subsection.

**Attack Goals**

An attacker always has specific goals. This study focuses on two potential attack goals: (i) create congestion for a specific approach at a subject intersection so that vehicles experience more waiting time; and (ii) fool a machine learning based predictive attack detection model, which predicts number of vehicles every second and ensures that there is no unusual sudden jump of number of vehicles to alter the traffic signal timing and phase.

*Congestion Creation*

An attacker may receive financial (or political) benefits to create a congestion for a specific roadway. The ultimate goal of the attacker will be blocking a route by generating fake vehicles at a slow rate so that it changes the signal time and phasing without any abrupt change in number of vehicles.

*Fooling a Machine Learning based Attack Detection Model*

As we generate congestion by creating fake vehicles, a recurrent neural network model can be used to detect such attack by predicting the number of vehicles on a specific approach at a subject intersection *(9)*. It will be challenging for a prediction-based detection model to identify an unusual vehicle arrival rate if an attacker slowly increases the number of fake vehicles that follow traffic and car-following rules. This slow increase can eventually fool the prediction model and thus render it unable to detect such attacks. For this purpose, we have introduced this new type of cyber attack—i.e., "slow poisoning"—in this paper.

**Simulation Set-up**

**Figure 2** shows screenshots from the SUMO *(16)* graphical user interface, which shows the geometry of nodes and edges. A node represents an intersection, and an edge refers to a roadway segment connecting the two nodes. Dedicated left-turn pockets are used at the junction for the left-turn movements. The connections define the possible vehicle movements that a vehicle can take when it arrives at a junction (see **Figure 2**). The connections are the same for all the junctions in the roadway network. Figure 2(a) presents the different components of the SUMO simulation environment. We have used the NETEDIT module of SUMO to define the vehicular demand. We have also created multiple flows (from-to) by selecting peripheral start and end edges for each flow. By default, the NETEDIT module uses the shortest minimum path between the start and end edge for the flow creation. We have used default vehicle types for the simulation. The simulation does not include any person or pedestrians.

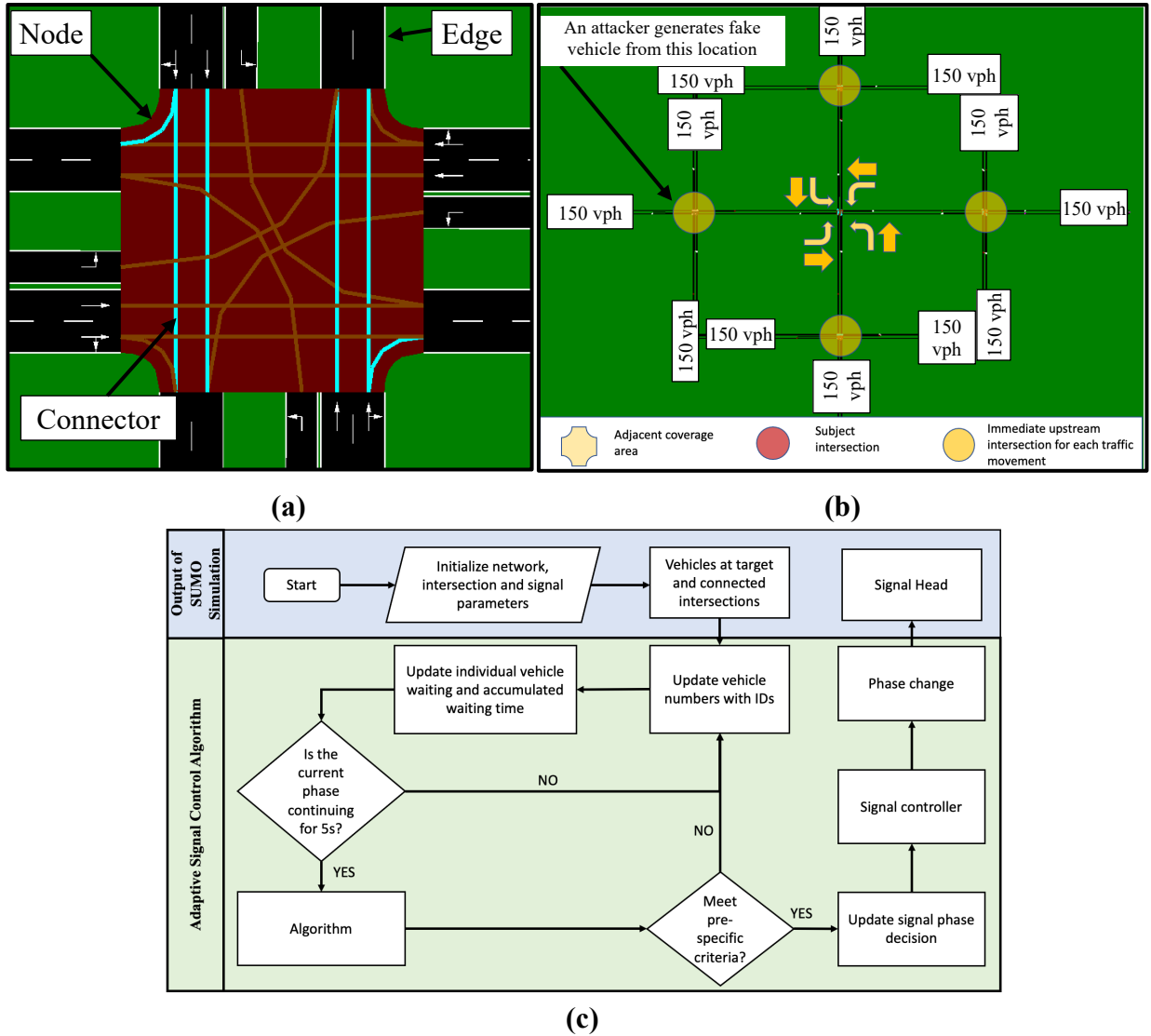

**Figure 2 Simulation set-up in SUMO for attack modeling and data generation: (a) intersection details; (b) network details; and (c) traffic signal control logic details**

**Waiting time-based Adaptive Traffic Signal Control Algorithm and Data Generation**

We have used Traffic Control Interface (TraCI) *(17)* for collecting vehicle level data and controlling the signal controller of the subject intersection in an adaptive manner. TraCI's inbuilt functions were used to collect vehicle ID and the waiting time for the edges that we define as approach waiting time (AWT).The waiting time based adaptive signal control algorithm allocates green time if traffic movement experiences the highest average waiting time out of eight movements (east bound through (EBT), east bound left (EBL), west bound through (WBT), west bound left (WBL), south bound through (SBT), south bound left (SBL), north bound through (NBT) and north bound left (NBL)). The vehicle is considered to incur wait time if its speed is 0.1 m/s or lower (as per the SUMO user guide *(17)*).

The approach average waiting time (*AAWT*) of *n* vehicles is derived using the following **equation 1.**

$$AAWT_t = \frac{AWT}{n} \; ; \tag{1}$$

where *AAWT* represents the average waiting time in seconds/vehicle, and *n* is the total number of vehicles that are approaching towards a subject intersection. *n* represents the number of vehicles on respective approaches for calculating *AAWT* for each approach and represents the total number of vehicles approaching the subject intersection while calculating the intersectional *AAWT*. The waiting time-based ATSC allocates green to a movement (through or left-turns) with the maximum *AAWT* at time *t* compared to other movements. Thus, the phase for green allocation is decided as:

$$max(AAWT_t^M); \; where, M = EBL, EBT, WBL\ WBT, NBL, NBT, SBL, and\ SBT$$

    We have implemented the waiting time-based algorithm in SUMO (see **Figure 2(c)**) and recorded the approach waiting time (AWT) of approach at the subject intersection every 1 second. The phase change decision is evaluated after every 5 seconds from the start of the green phase. A phase change, when warranted, follows through a yellow change interval of 2 seconds and 1 second of all red clearance. A phase change is not warranted if the approach is currently green and still has maximum AWT after the 5 seconds from the start of the last green. The yellow of 2 seconds and red of 1 second is skipped in such cases. The traffic head is controlled based on the algorithm output. All the simulation runs were made for a total of 3,600 seconds including the 600 seconds of warm-up period and another 600 seconds of cool-down period. The results reported henceforth relate to the 2,400 seconds of actual simulation period only.

    We have generated fake vehicles if there is an enough gap between actual vehicles following traffic and car-following rules from starting edge of the EB approach of the subject intersection (see **Figure 1**). Using a trial-and-error approach, different traffic demands have been tested by changing the vehicle generation per hour (vph) in the flow attributes in SUMO to find an optimum demand for a "slow poisoning" attack. Note that, the traffic demand in the attack free simulation scenario from each entry point of simulated network is 150 vph (see **Figure 2(b)**). **Figure 3 (a, b, c, and d)** presents profiles of number of vehicles and average waiting time of the approaching vehicles towards the subject intersection in the EB approach without an attack. As our waiting time-based ATSC algorithm uses AAWT to regulate the signal phase, a prediction model can be trained with these data and predict the number of vehicles for the next time stamp. In this way, any anomalies in predicting number of vehicles can lead to attack detection. In this approach, an attack can be detected using a threshold, which is the maximum error that can be generated by a prediction model.

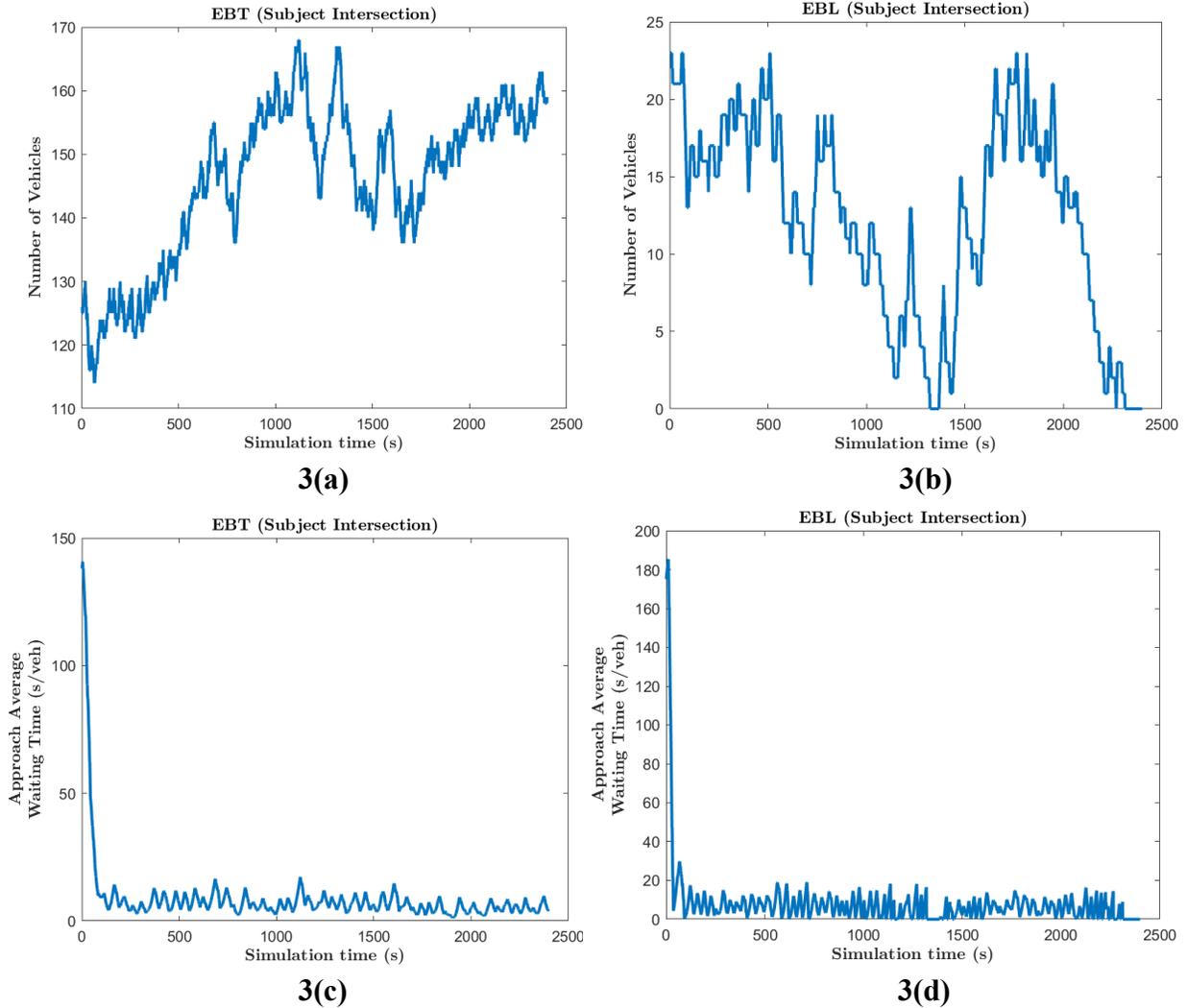

**Figure 3** Average waiting time and number of vehicles at the subject intersection (EB approach only)

**Justifications for Achieving Attack Goals**

In this subjection, we justify how an attacker achieves his/her attack goals—i.e., congestion creation and fooling machine learning based attack detection—and create a sophisticated "slow poisoning" attack.

*Congestion Creation*

**Figure 4** presents an example of before and after attack scenarios, which is simulated in SUMO. When an attacker starts creating fake vehicles at the starting point of the EB approach of the subject intersection, it lowers the average waiting time of the approach. Thus, the ATSC algorithm gives green phase to the other approaches that results in queue building up in the EB approach. As the attacker continuously generates fake vehicles to keep the AAWT lower than other vehicles, the congestion situation becomes worse. Note that it will be a challenge for the attacker to make a tradeoff between increasing number of vehicles and rules of traffic and car-following behaviour.

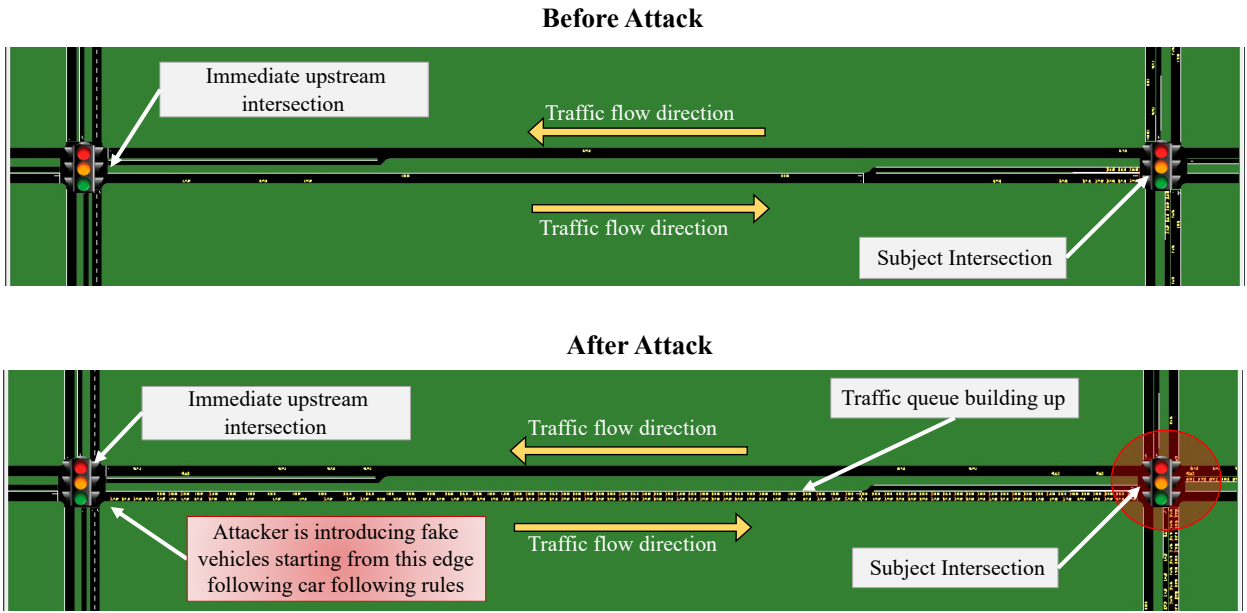

**Figure 4 An example before and after attack scenarios using the SUMO microscopic traffic simulator**

**Figure 5** presents attack free and attack data in terms of number of vehicles. We have presented two separate profiles as our attack free scenario is a separate simulation run. In the attack scenario, we have generated fake vehicles at a slow rate to generate a "slow poisoning" attack on the waiting time-based ATSC. As shown in **Figure 5**, we started an attack at 400 seconds. After 400 seconds, we observe that the number of vehicles in the EB approach increases. However, around 1000 seconds, the number of vehicles decreases again as our waiting time-based ATSC allocates green time depending on the AAWT. Thus, a queue builds between 400 seconds and 1000 seconds. After that, we observe that there are five different time slots (see light blue shaded area on Figure 5) where the number of vehicles increases in the EB approach. Note that an attacker must be creative how he/she can generate a certain increase of vehicles to build a congestion. At the same time, the ATSC keeps regulating green phase as the waiting time increases with increasing the number of vehicles. Thus, the number of vehcles decreases after 1,000 seconds again. However, if the number of vehicle increases, it could decrease the approach average waiting. Because of the dynamic nature of vehicle arrival rate and ATSC, it is very hard to create and detect this type of attack, which we introduce as a "slow poisoning" attack for a waiting time-based ATSC.

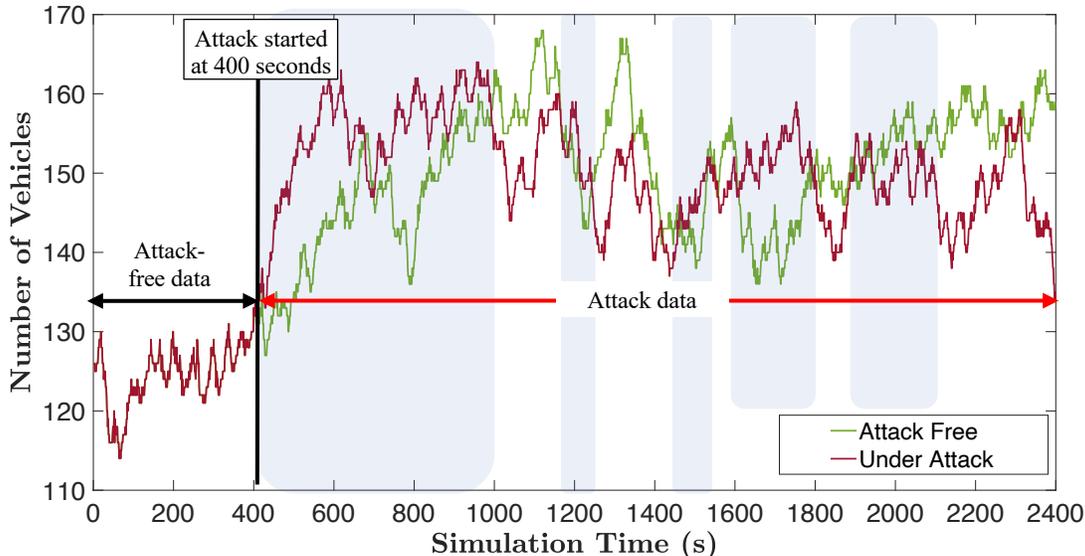

**Figure 5 Attack free and under attack dataset EB approach combining EBT and EBL approach**

*Fooling Machine Learning based Attack Detection*

An attack can be detected if we predict the arrival rate of vehicles for the specific approach of a subject intersection. However, it is challenging to predict the number of vehicles because of stochastic arrival rate, which depends on the time of the day, day of the week, and weather. However, existing literature shows that the number of vehicles can be predicted with high accuracy using a recurrent neural network *(2)*. Specifically, the long-short term memory (LSTM) model can capture the relationship in a time series data and predict traffic flow with high accuracy in a connected vehicle environment. In this subsection, we establish that an LSTM model follows the pattern of attack dataset because of the following reasons: (i) the sophisticated nature of attack— i.e., "slow poisoning"; and (ii) the characteristics of ATSC. We have established a threshold based on the prediction model maximum absolute error (see **Equation (2)**). If the difference between the perceived number of vehicles and the predicted number of vehicles is greater than the error threshold, an attack will be detected. Note that, for detecting an attack, we are capitalizing on the model's inability to predict the number of vehicles accurately.

$$Error\ Threshold = Prediction\ Model\ Maximum\ Absolute\ Error \qquad (2)$$

We have predicted the number of vehicles using a 2-stacked LSTM (28) with 128 and 8 neurons in the first and second hidden layers, respectively. The training and validation data includes the number of vehicles and waiting time, as shown in **Figure 3**. The output is the predicted number of vehicles. In this study, the data generation frequency is 1s, which is the update time of our ATSC. The dataset is split into training with 70% and validation with 30% observations of 2400 seconds simulation time. Before feeding the number of vehicles and waiting time into the LSTM for training, the input features are normalized between 0 and 1. We have used a trial-and-error approach for determining the LSTM hyperparameters—i.e., number of neurons, number of epochs, batch size, and learning rate—as it is a time series-based prediction model. The Mean Absolute Error (MAE) metric is used as the loss function for evaluating overfitting and underfitting while training the model. To evaluate the goodness-of-fit of the LSTM model, the loss profiles using optimal hyperparameters are shown in **Figure 6(a)** with both the training and validation datasets. A comparison of the MAE of these two datasets indicates a good fit of the prediction model with the optimal hyperparameters. The values of hyperparameters and name of the optimizer are listed in **Table 1**. After validating the prediction model, we found that the Root Mean Square Error (RMSE) for the predicted number of vehicles is 1 vehicle (rounded up from 0.58 to 1 as the vehicles number cannot be a fractional value); and the maximum absolute error is 3 vehicles (rounded up from 2.09 to 3). **Figure 6(b)** presents the validation outcome, which is the comparison between the predicted and ground truth data.

**TABLE 1 Optimal Hyperparameters of LSTM Model**

| Hyperparameters and Optimizer | Value |
|---|---|
| Number of neurons (1$^{st}$ layer) | 128 |
| Number of neurons (2$^{nd}$ layer) | 8 |
| Number of epochs | 1000 |
| Batch size | 50 |
| Learning rate | 0.01 |
| Optimizer | adam |

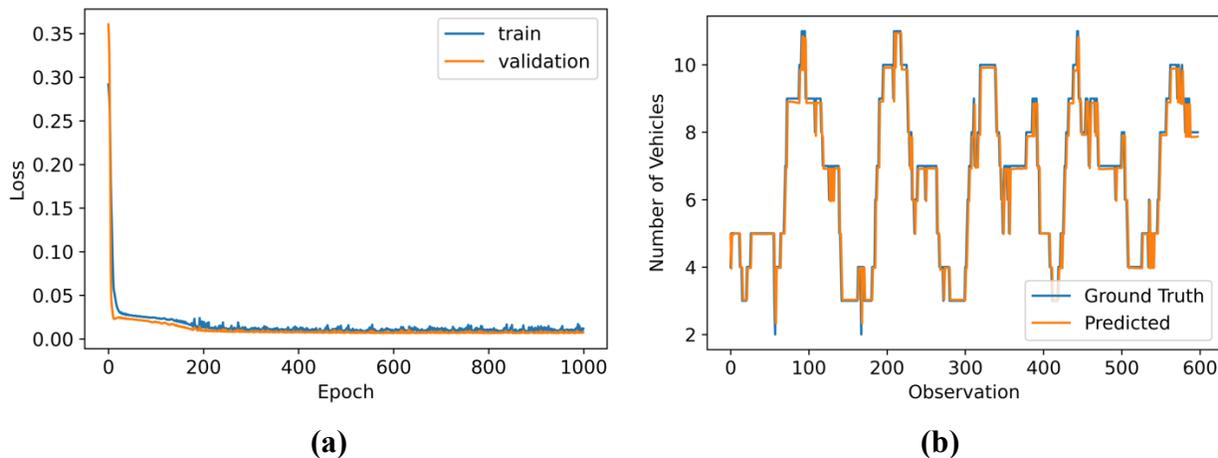

(a)        (b)

**Figure 6 (a) Comparison of Mean Absolute Error (MAE) (loss) profiles with the optimal parameter set; and (b) Ground truth and predicted number of vehicle profiles for the validation dataset.**

**Figure 7** presents the compromised and predicted profiles of the number of vehicles for the EB approach of the subject intersection. Although the attack is created at 400 seconds, we observe that the predicted profile follows the compromised data very closely. As shown in **Figure 8**, at around 1,750 seconds, the prediction model crosses the error threshold. Before 1750 seconds, the prediction model could not detect any attack at all. Note that the queue builds up after 1,800 seconds if we compare with the number of vehicles in the attack free data. A possible explanation for this is that the adaptive traffic signal control algorithm allocates more green time as queue is building up and waiting time is increasing. Thus, it justifies that our innovative "slow poisoning" attack could fool the prediction-based attack detection system.

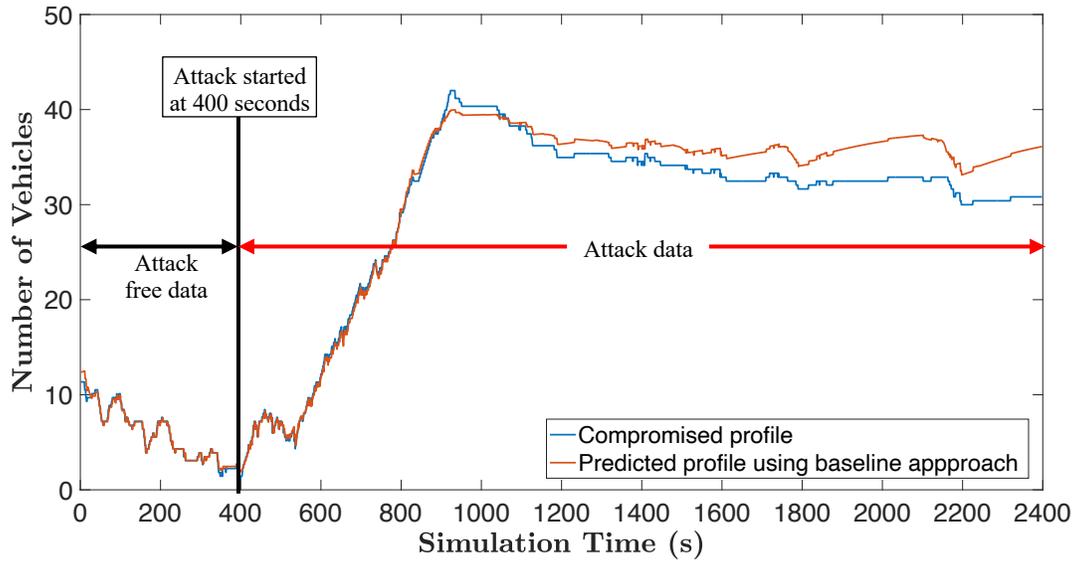

**Figure 7** Compromised and predicted profiles of number of vehicles for the subject intersection (2400 seconds simulation period).

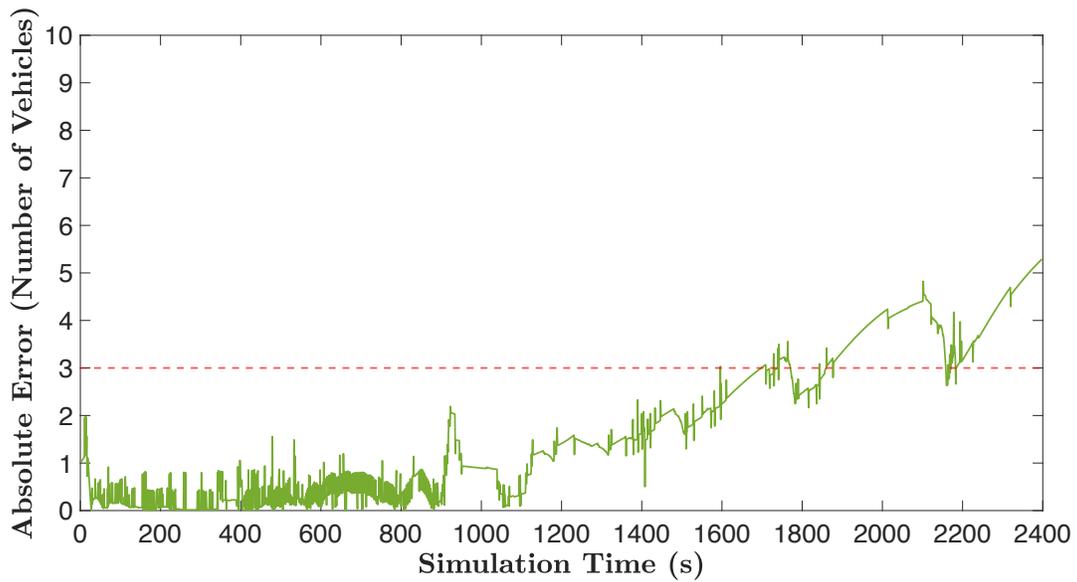

**Figure 8** Absolute prediction error profiles using the baseline approach.

# DEVELOPMENT OF AN ATTACK DETECTION MODEL AND ANALYSES OUTCOME

In this section, we have developed a detection strategy for a "slow poisoning" attack on a waiting time-based ATSC algorithm. The details of the detection strategy are shown in **Figure 9**. Although we use the same architecture of the LSTM model, our strategy is different for "slow poisoning" attack detection. Instead of only using number of vehicles and average waiting time of the EB approach, we use number of vehicles and corresponding waiting time for those vehicles approaching the subject intersection from the immediate upstream intersection (North bound left, South bound right, and East bound through). Overall, our feature set is different compared to the baseline strategy (as shown in the previous section) for predicting number of vehicles for an approach. As an attacker is creating fake vehicles at the starting point of EB lanes of the subject intersection, we assume the vehicles that are coming towards the immediate upstream intersection will not be affected by the attacker. As before, we have established a threshold based on the prediction model maximum absolute error. If the difference between the actual number of vehicles and the predicted number of vehicles is greater than the error threshold, an attack will be detected. Again, we are capitalizing the model's inability to predict the number of vehicles accurately for detecting an attack.

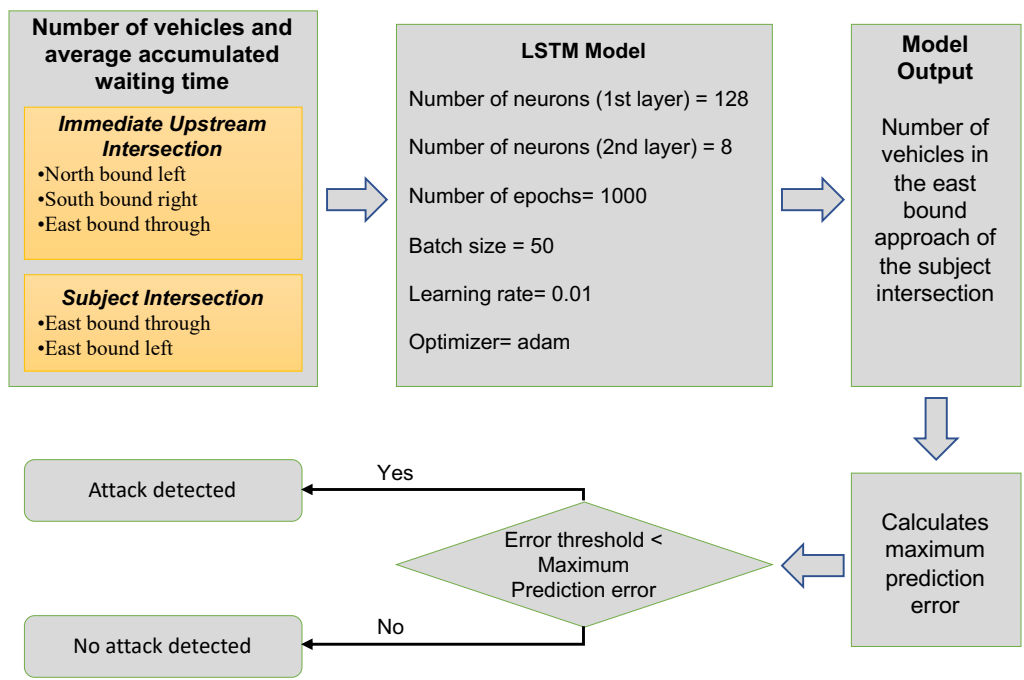

**Figure 9 "Slow poisoning" attack detection strategy**

As shown in **Figure 9**, we use the same 2-stacked LSTM (28) with 128 and 8 neurons in the first and second hidden layers, respectively. The training and validation data includes the number of vehicles and waiting time for the EB approach and the vehicles that are coming towards the EB approach form the immediate upstream intersection. The output from the LSTM will be the number of vehicles. As before, the data generation frequency is 1s, which matches with the update time of our ATSC. As mentioned before, ee have split the dataset into training with 70% and validation with 30% observations of 2400 observations. Before feeding the number of vehicles and waiting time to the LSTM training, we have also normalized all data between 0 and 1.

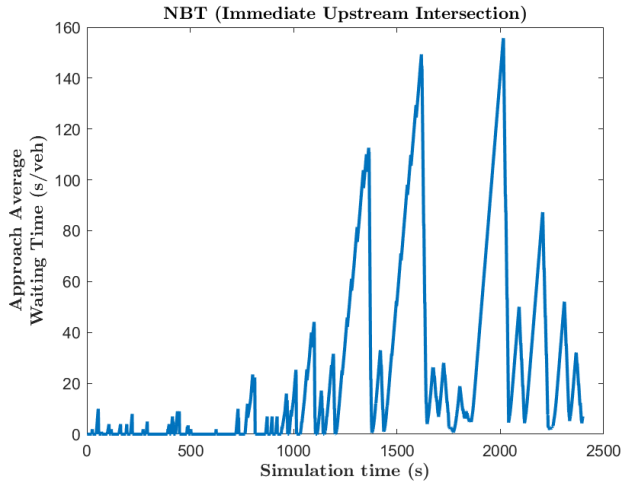
(a)

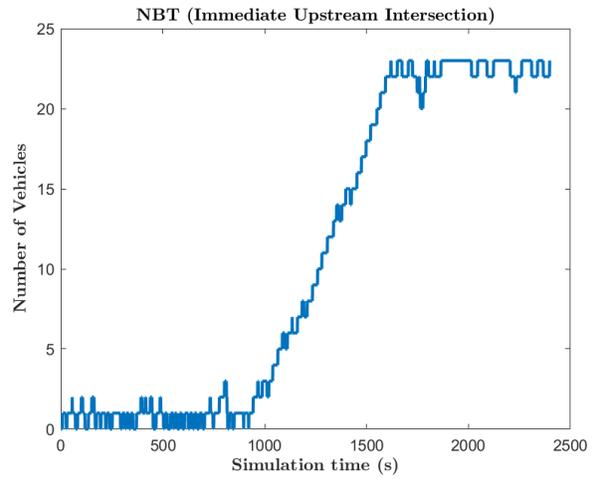
(b)

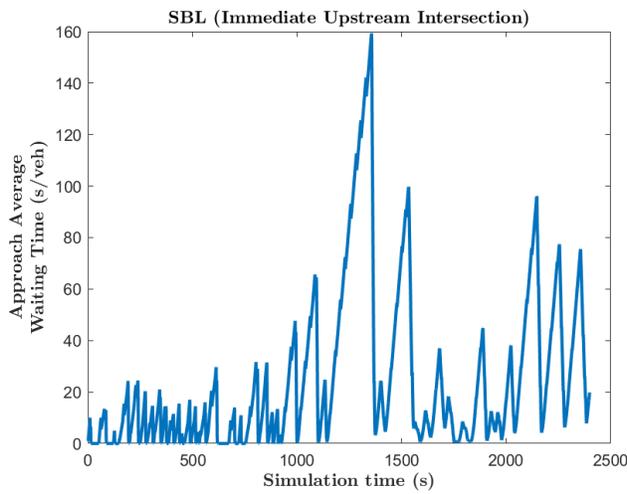
(c)

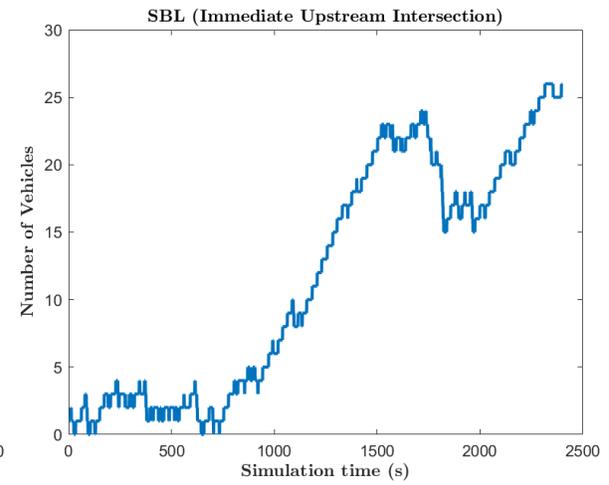
(d)

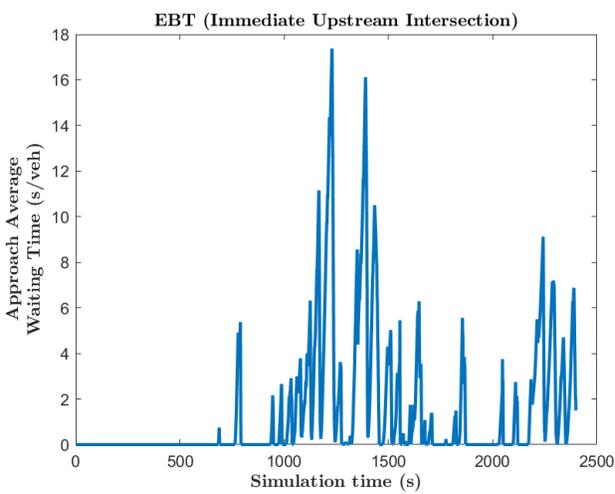
(e)

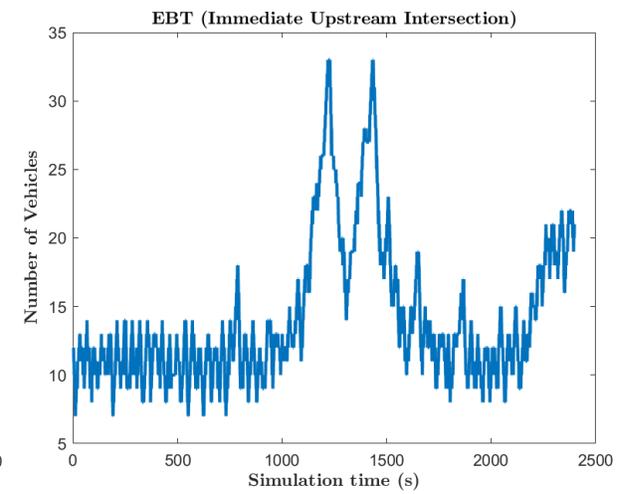
(f)

**Figure 10 Waiting time and number of vehicles at the immediate upstream intersection**

Again, we used a trial-and-error approach for determining the LSTM hyperparameters—i.e., number of neurons, number of epochs, batch size, and learning rate—as it is a time series-based prediction model. The Mean Absolute Error (MAE) metric is used as the loss function for evaluating overfitting and underfitting while training the model. To evaluate the goodness-of-fit of the LSTM model, the loss profiles using optimal hyperparameters are shown in **Figure 11 (a)** with both the training and validation datasets. A comparison of the MAE of these two datasets indicates a good fit of the prediction model with the optimal hyperparameters. The values of hyperparameters and name of the optimizer are listed in **Table 1 and Figure 9**. After validating the prediction model, we found that the Root Mean Square Error (RMSE) for the predicted number of vehicles is 1 vehicle (rounded up from 0.56 to 1), and the maximum absolute error is 3 vehicles (rounded up from 2.25 to 3). **Figure 11(b)** presents the validation outcome, which is the comparison between the predicted and ground truth data.

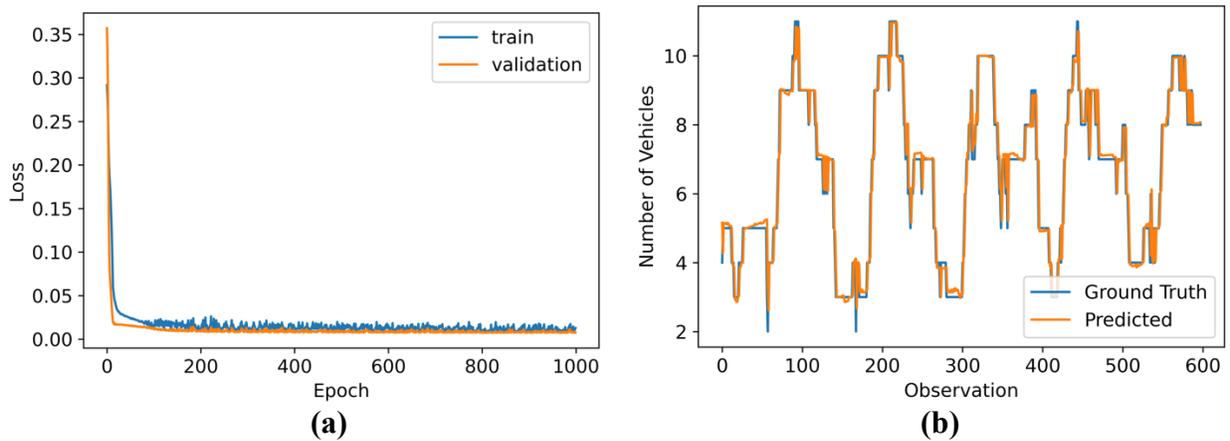

(a)  (b)

**Figure 11 (a) Comparison of Mean Absolute Error (MAE) (loss) profiles with the optimal parameter set; and (b) Ground truth and predicted number of vehicle profiles using the validation dataset.**

**Figure 12** presents a comparison between datasets, prediction results, and prediction error profiles using the proposed detection approach. The light blue shaded area shows where the maximum prediction error crosses the error threshold. The number of vehicles increases in those shaded areas and our detection strategy is able to flag an attack. We see that the number of vehicles decreases at some locations in the attack dataset compared to the attack-free dataset. A possible explanation for this is that the adaptive traffic signal control algorithm allocates more green time as the queue is building up and waiting time is increasing. Thus, many vehicles leave the intersection and thereby reduce the number of vehicles after this increase. The number of vehicles again increases when the attacker injects more fake vehicles that follow the rules of traffic and car-following. This proves that our attack detection algorithm is able to flag an attack.

*Dasgupta, Hollis, Rahman, and Atkison*

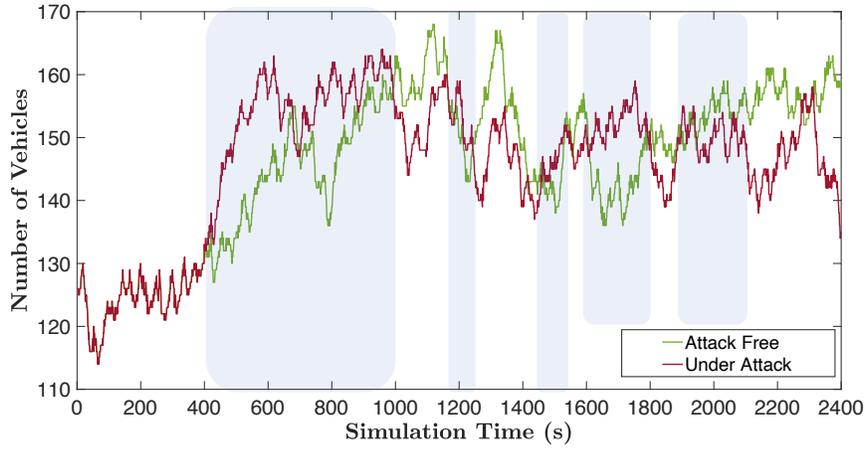

(a)

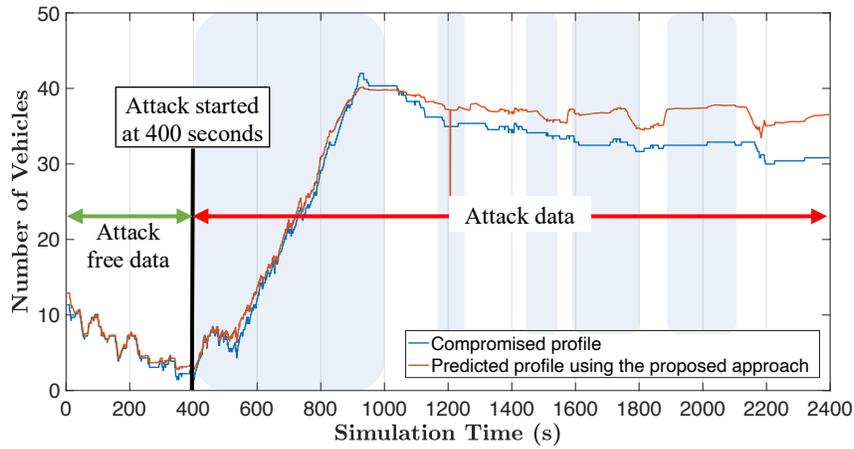

(b)

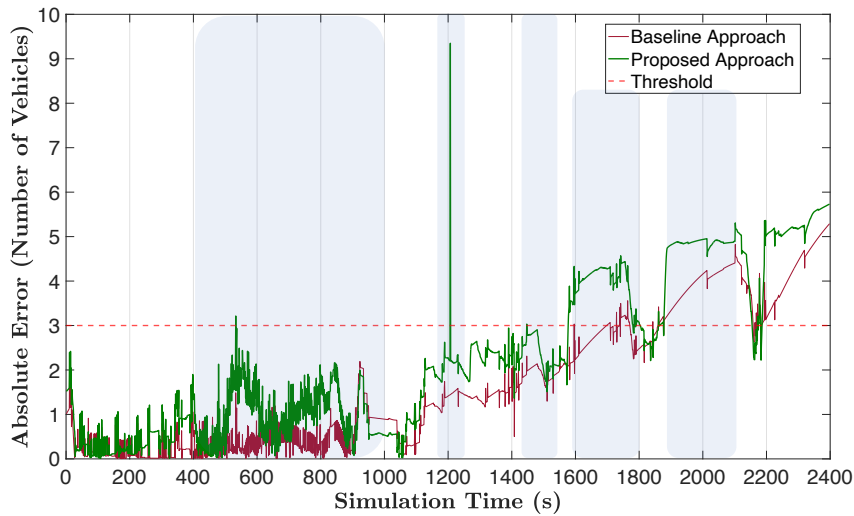

(c)

**Figure 12 Comparison between (see the light blue shaded area): (a) attack free and under attack dataset; (b) compromised and predicted profiles using the proposed approach; and (b) absolute prediction error profiles using the baseline and the proposed approach.**

# CONCLUSIONS

This paper introduces an innovative "slow poisoning" cyberattack for a waiting time based ATSC and corresponding detection strategy. An attacker can block a route by generating fake vehicles following traffic and car-following rules at a slow rate so that it changes the signal timing and phase without having any abrupt changes in the number of vehicles. It is difficult to model this kind of attack because of the adaptive nature of ATSC and the stochastic nature of vehicle arrival rate at the intersection; it is also challenging to develop a strategy to detect such attack. We have developed a "slow poisoning" attack generation strategy for an adaptive traffic signal controller and a prediction-based "slow poisoning" attack detection strategy using a recurrent neural network—LSTM model. We have also generated a "slow poisoning" attack modeling strategy using a microscopic traffic simulator—SUMO—and used generated data from the simulation to develop the attack detection model. Our analyses revealed that our detection strategy is able to flag an attack if an attacker tries to generate fake vehicles and block an approach. In our follow-up study, we will develop a mathematical model of a "slow poisoning" attack and will prove the efficacy of our strategy in different traffic demand scenarios.

## ACKNOWLEDGMENTS

This material is based on a study partially supported by seed grant by the University of Alabama Cyber Initiative. Matthew Shrode Hargis of the University of Alabama provided valuable assistance in editing the paper.

## AUTHOR CONTRIBUTIONS

The authors confirm contribution to the paper as follows: study conception and design: S. Dasgupta, M. Rahman; data collection: S. Dasgupta and M. Rahman; interpretation of results: S. Dasgupta, M. Rahman, C. Hollis, and T. Atkison; draft manuscript preparation: S. Dasgupta, M. Rahman, C. Hollis, and T. Atkison. All authors reviewed the results and approved the final version of the manuscript.